\documentclass[11pt, letterpaper, twocolumn]{article}

\usepackage{amsmath,amssymb,amsfonts}
\usepackage{algorithm}
\usepackage{algpseudocode}
\usepackage{algorithmicx}
\usepackage{graphicx}
\usepackage{textcomp}
\usepackage{xcolor}
\usepackage{microtype}
\usepackage{graphicx}
\usepackage{amssymb}
\usepackage{epstopdf}
\usepackage{booktabs}
\usepackage{multirow}
\usepackage{threeparttable}
\usepackage{subfigure}
\usepackage{float}
\usepackage{amsmath}
\usepackage{setspace}
\usepackage{bm}
\usepackage{hyperref}
\usepackage{cite}
\def\BibTeX{{\rm B\kern-.05em{\sc i\kern-.025em b}\kern-.08em
    T\kern-.1667em\lower.7ex\hbox{E}\kern-.125emX}}
\usepackage{authblk}

\let\hat\widehat

\newcommand{\bl}{\bm{l}}

\newcommand{\bt}{\bm{t}}

\newcommand{\bx}{\bm{x}}
\newcommand{\by}{\bm{y}}
\newcommand{\bz}{\bm{z}}

\newcommand{\bC}{\bm{C}}

\newcommand{\bG}{\bm{G}}
\newcommand{\bH}{\bm{H}}
\newcommand{\bI}{\bm{I}}
\newcommand{\bJ}{\bm{J}}
\newcommand{\bK}{\bm{K}}

\newcommand{\bX}{\bm{X}}

\newcommand{\bZ}{\bm{Z}}

\newcommand{\cA}{\mathcal{A}}

\newcommand{\cC}{\mathcal{C}}

\newcommand{\cH}{\mathcal{H}}

\newcommand{\cT}{{\mathcal{T}}}
\newcommand{\cU}{\mathcal{U}}
\newcommand{\cV}{\mathcal{V}}

\newcommand{\cX}{U}
\newcommand{\cY}{V}
\newcommand{\cZ}{W}
\def\ci{\perp\!\!\!\!\perp}
\def\nci{\not\!\perp\!\!\!\perp}

\newcommand{\EE}{\mathbb{E}}

\newcommand{\RR}{\mathbb{R}}

\newcommand{\bbeta}{\bm{\beta}}

\newcommand{\Epsilon}{\bm{\Epsilon}}

\newcommand{\bOmega}{\bm{\Omega}}

\newcommand{\diag}{{\rm diag}}

\newcommand{\inner}[2]{\left\langle #1,#2 \right\rangle}

\newcommand{\bbE}{\mathbb{E}}

\newcommand{\bbR}{\mathbb{R}}

\numberwithin{equation}{section}
\newtheorem{thm}{Theorem}[section]

\newtheorem{lem}{Lemma}[section]

\begin{document}
	
\title{Neural Gaussian Mirror \\for Controlled Feature Selection in Neural Networks}

\author[1]{Xin Xing\thanks{The first two authors contributed equally to this paper.}}
\author[2]{Yu Gui\thanks{The first two authors contributed equally to this paper.}}
\author[3]{Chenguang Dai}
\author[3]{Jun S. Liu}
\affil[1]{Department of Statistics, Virginia Tech}
\affil[2]{Department of Statistics, University of Chicago}
\affil[3]{Department of Statistics, Harvard University}
\date{}
\maketitle

\begin{abstract}
Deep neural networks (DNNs) have become increasingly popular and achieved outstanding performance in predictive tasks. However, the  DNN framework itself cannot inform the user which features are more or less relevant  for making the prediction, which limits its applicability in many scientific fields. We introduce neural Gaussian mirrors (NGMs), in which  mirrored features are created, via a structured perturbation based on a kernel-based conditional dependence measure, to help evaluate feature importance. We design two modifications of the DNN architecture for incorporating mirrored features and providing mirror statistics to measure feature importance. As shown in simulated and real data examples, the proposed method controls the feature selection error rate at a predefined level and maintains a high selection power even with the presence of highly correlated features.
\end{abstract}

\section{Introduction}
Recent advances in deep neural networks (DNNs) significantly improve the state-of-the-art prediction methods in many fields such as image processing, healthcare, genomics, and finance. However,  complex structures of  DNNs often lead to their lacks of interpretability. In many fields of science, model interpretation is essential for understand the underlying scientific mechanisms and for rational decision-making.  At the very least, it is important to know which features are more relevant to the prediction outcome and which are not, i.e., conducting feature selection.
Without properly assessing and controlling the reliability of such feature selection efforts, the reproducibility and generalizability of the discoveries are likely  unstable. For example, in genomics, current developments lead to good predictions of complex traits using millions of genetic and phenotypic predictors. If one trait is predicted as ``abnormal,'' it is important for the doctor to know  which predictors are active in such a prediction, whether they make medical and biological sense, and whether there exist targeted treatment strategies. 

In the classical statistical inference, some $p$-value based methods such as the Benjamini-Hochberg  \cite{benjamini1995controlling} and Benjamini-Yekutieli  \cite{benjamini2001control} procedures  have been widely employed for controlling the false discovery rate (FDR). However, in nonlinear and complex models, it is often difficult to obtain valid p-values since the analytical form of the distribution of a relevant test statistics is largely unknown. Approximating  p-values by bootstrapping and other resampling methods is a valid strategy, but  is also often infeasible due to model non-identifiability, (i.e., the existence of multiple equivalent models, such as a DNN with permuted input or internal nodes), complex data and model structures, and high computational costs. 
In an effect to code with these difficulties,
\cite{barber2015controlling} proposed the knockoff filter, which constructs a ``fake" copy of the feature matrix that is independent of the response but retains the covariance structure of the original feature matrix (also called the ``design matrix").
Important features can be selected after comparing inferential results for the true features with those for the knockoffs. To extend the idea to higher dimensional and more complex models,
\cite{candes2018panning} further proposed the Model-X knockoff, which considers a random design with known joint distribution (Gaussian) of the input features.  
Recently, \cite{yu2019three} develops the PCS inference to investigate the stability of the data by introducing perturbations to the data, the model, and the algorithm.  Dropout \cite{wager2013dropout} and Gaussian Dropout can be viewed as examples of model perturbations, which are efficient for reducing the instability and elevating out-of-sample performances. 

For enhancing interpretability, feature selection in neural networks has attracted some recent attention.
To this end, gradients with respect to each input are usually taken as an effective measurement of feature importance in convolutional and multi-layer neural networks as  shown in \cite{hechtlinger2016interpretation}.
\cite{verikas2002feature} selects features in DNNs based on changes of the cross-validation (CV) classification error rate  caused by the removal of an individual feature. However, how to control the error of such a feature selection process in DNNs is unclear.
A recent work by \cite{lu2018deeppink} utilizes the Model-X Knockoff method to construct a pairwisely-connected input layer to replace the original one, and the latent ``competition'' within each pair separates  important features from unimportant ones by the measurement of knockoff statistics. However, the Model-X method requires that the distribution of $\bX$ be either known exactly, or known to be in a nice distribution family that possesses simple sufficient statistics  \cite{huang2019relaxing}.
Also, the construction procedure for the knockoffs in  \cite{lu2018deeppink} is based on the Gaussian assumption on the input features, which is not satisfied in many applications such as genome wide association studies with discrete input data.
In addition, the DeepPINK does not fit exactly the purpose of network interpretation since the input layer is not fully connected to hidden layers, and thus some of the connections in the ``original'' structure may be lost in this specific Pairwise-Input structure.

In this paper,  we introduce the neural Gaussian mirror (NGM) strategy for feature selection with controlled   false selection error rates. NGM does not require any  knowledge about or assumption on the joint distribution of the input features. Each NGM is created by perturbing an input feature explicitly.   For example, for the $j$th feature, we create two mirrored features as $\bX_j^+ = \bX_j + c_j \bZ_j$ and $\bX_j^-=\bX_j - c_j \bZ_j$, where  $\bZ_j$ is the vector of \textit{i.i.d.} standard Gaussian random variables. The scalar $c_j$ is  chosen so as to minimize the dependence between $\bX_j^+$ and $\bX_j^-$ conditional on the remaining variables, $\bX_{-j}$. To cope with  nonlinear dependence, we introduce a kernel-based conditional independence measure and obtain $c_j$ by solving an optimization problem. Then, we construct two modifications of the network architectures based on the mirrored design. %
The proposed mirror statistics tend to take large positive values for  important features and are symmetric around zero for null features, which enables us to control the selection error rate.

\section{Background}

\subsection{Controlled feature selection }

In many modern applications, we 
are not only interested in fitting a model (with potentially many predictors) that can achieve a high prediction accuracy, but also keen in selecting relevant features with controlled false selection error rate.
Mathematically, we consider a general model with an unknown link function $F$ illustrating the connection between the response variable $Y$ and $p$ predictor variables $X\in \mathbb{R}^{p}$: $Y=F(X)+\epsilon$, where $\epsilon$ denotes the random noise.

We define $S_0$ to be the set of ``null" features:  $\forall \ k\in S_0$, we have  $Y \ci X_j \mid  X_{-j}$,
where $X_{-j}= \{X_1,\dots, X_p\}\backslash \{X_j\}$
and define $S_1 = \{1,\dots,p\}\backslash S_0$ as the set for relevant features. 
Feature selection is equivalent to recovering $S_1$ based on observations. When the estimated set $\hat{S}_1$ is produced, the false discoveries can be denoted as $\hat{S}_1 \cap S_{1}^{c}$, where $S_1$ is the true set of important features. The {\it false discovery proportion} (FDP) is defined as
\begin{equation}\label{eq:fdp}
\text{FDP} = \frac{|\hat{S}_1 \cap S_{1}^{c}|}{|\hat{S}_1|},
\end{equation}
and its expectation is called the {\it false discovery rate} (FDR), i.e., $ {\rm FDR} = \mathbb{E}[{\rm FDP}]$.

Technically, controlled feature selection methods are different from conventional feature selection methods such as Lasso  in that the former require additional efforts in  assessing the selection uncertainty by means of estimating the FDP, which
becomes more challenging in DNNs.

\subsection{Gaussian mirror design for linear models}
To  discern  relevant features from irrelevant ones, we intentionally perturb the input features via the following {\it Gaussian mirror design}: for each feature $X_j$, we construct a mirrored pair $(X_j^{+},X_j^{-}) = (X_j+c_j Z_j,X_j-c_j Z_j)$ with $Z_j\sim N(0,1)$ being independent of all the $X$. We call $(X_j^+, X_j^-, X_{-j})$ the $j$-th  mirror design. Regressing $Y$ on the $j$-th mirror design, we  obtain coefficients  via the ordinary least squares (OLS) method. Let $\hat{\beta}_j^{+}$ and $\hat{\beta}_j^{-}$ be the estimated coefficients for $X_j^+$ and $X_j^-$, respectively. The mirror statistics for the $j$th feature is defined as:
\begin{figure}[!t]
    \centering 
	\includegraphics[width = 0.5\textwidth]{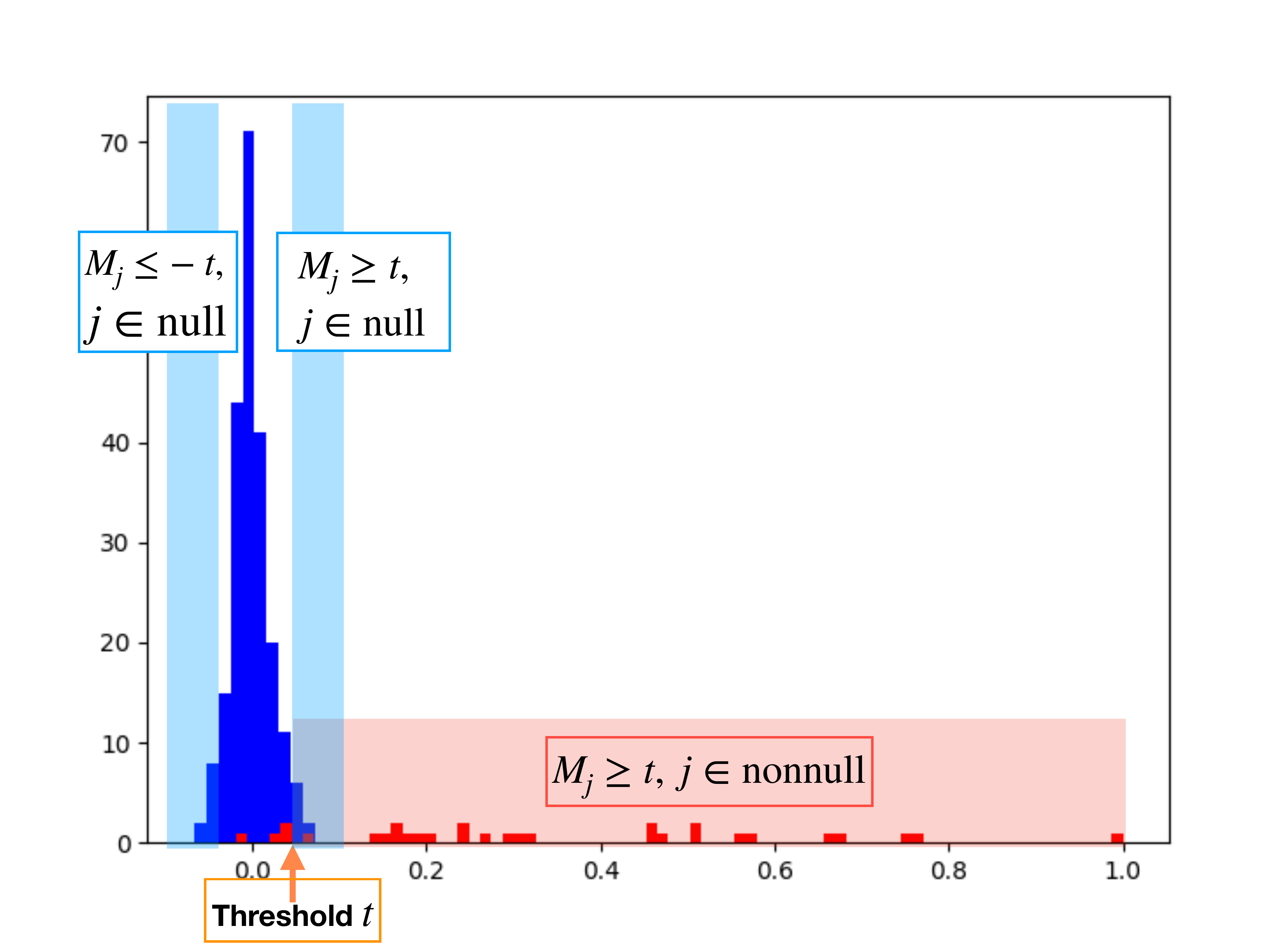}
	\caption{Histogram of mirror statistics: red represents important variables and blue stands for null variables}
	\label{fig:ms}
\end{figure}
\begin{align}
M_j &= |\hat{\beta}_j^{+}+\hat{\beta}_j^{-}|-|\hat{\beta}_j^{+}-\hat{\beta}_j^{-}|
\end{align}
For important features $j\in S_1$, $\hat{\beta}_j^{+}$ tends to be similar to  $\hat{\beta}_j^{-}$, which helps cancel out  the perturbation part leading to a large positive value of $M_j$. From another point of view, the large value of $M_j$ indicates that the importance of the $j$th features is more stable to the perturbation, i.e. $\hat{\beta}_j^{+}\approx \hat{\beta}_j^{-}$.  For null features $j\in S_0$, the mirror statistics can be made symmetric about zero by choosing a proper  $c_j$ as detailed in Theorem~\ref{thm:gmlinear}. 

\begin{thm}\label{thm:gmlinear}
	Assume $Y= X\beta + \epsilon$ where $\epsilon$ is Gaussian white noise. Let $X_j^+= X_j+cZ_j$, $X_j^- = X_j-c Z_j$, and $I^{L}_j(c) = \rho_{X_j^+ X_j^-\cdot X_{-j}}$, i.e., the partial correlation of $X_j^+$ and $X_j^-$ given $X_{-j}$. For any $j \in S_0$, if we set  
	\begin{equation}\label{eq:cjlm}
	c_j = \arg\min_{c} [I^{L}_j(c)]^2,
	\end{equation}
	then we have $P(M_j<-t) = P(M_j>t)$ for any $t>0$.
\end{thm}
A detailed proof of the theorem can be found in \cite{xing2019gm}.
It shows that the symmetric property of the mirror statistics $M_j$ for null features is satisfied if we choose $c_j$ that minimizes the magnitude of the partial correlation $I^{L}_j (c)$, which is equivalent to solving $I^{L}_j(c)=0$ in linear models. %
In other words, the perturbation $c_j Z_j$ makes $X_j^+$ and $X_j^-$ partially uncorrelated given $X_{-j}$.  

We note that in linear models with OLS fitting, the above construct is equivalent to generating an independent noise feature $c_j Z$ and regress $Y$ on $(\bX,c_jZ)$. The mirror statistic is simply the magnitude difference between the estimated coefficient for $X_j$ and that for $c_jZ$. However,  %
this simpler construct is no longer equivalent to the original design  for nonlinear models or for high-dimensional linear models where the OLS is not used for fitting. Also, the conditional independence motivation for choosing $c_j$ is lost in this simpler formulation.

Figure \ref{fig:ms} shows the distribution of the $M_j$'s. It is seen that relevant features can be separated from the null ones fairly well and  a consistent estimate of the false discoveries proportion based on the symmetric property can be obtained.

For a threshold $t$, we select features as
$\hat{S}_1 = \{j: M_j \geq t\}$.
The symmetric property of $M_j$ for $j\in S_0$ implies that $\bbE[\#\{j \in S_0 :\;M_j \geq t\}] = \bbE [\#\{j \in S_0:\;M_j \leq -t\}]$. Thus we obtain an estimate of the FDP as
\begin{align}\label{eq:fdphat}
\widehat{FDP}(t) = \frac{\#\{j :M_j \leq -t\}}{\#\{j : M_j \geq t\}\vee 1}
\end{align}
In practice, we choose a data adaptive threshold 
\begin{align}
\tau_q = \min_{t}\{t>0 :\widehat{FDP}(t) \leq q\}
\end{align} to control the FDR at a predefined level $q$.

As shown in Theorem 4 of \cite{xing2019gm}, under weak dependence assumption of $M_j$s, we have
\begin{equation*}
\EE[FDP(\tau_q)] < q
\end{equation*}
as $p$ goes to infinity, i.e., the FDR is asymptotically controlled under the predefined threshold $q$.

\section{Neural Gaussian mirror}

We here describe  a model-free mirror design for controlling variable selection errors in DNNs.  
In linear models, a key step of our mirror design is to choose a proper perturbation level $c_j$ so as to annihilate the partial correlation between the mirror variables $X_j^+$ and $X_j^-$. Since the partial correlation being zero for two random variables implies that they are  conditional independence under the joint Gaussian assumption, for more complex models such as DNNs, we consider a general {\it kernel-based} measure of conditional dependence between $X_j^+$ and $X_j^-$
and choose $c_j$ by minimizing this measure. 
Then, we can construct the mirror statistics and  establish a data-adaptive threshold to control the selection error rate in a similar way as in linear models. We call this whole procedure the neural Gaussian mirror (NGM).

\subsection{Model-free mirror design}
Without any specific model assumption, we construct the mirrored pair as $$(\bX_j^{+},\bX_j^{-})=(\bX_j+c_j \bZ_j,\bX_j-c_j \bZ_j),$$ 
where $\bZ_j\sim N(0, I_n)$, and $c_j$ is chosen as
\begin{align}
c_j &= {\rm arg}\min_{c}  [I^{K}_j(c)]^2
\end{align}
Here $[I_j^K(c)]^2$ is a kernel-based conditional independence measure between $X_j^+$ and $X_j^-$ given $X_{-j}$. We give a detailed expression of $[I_j^K(c)]^2$ in Sections 3.2 and 3.3. Compared with (\ref{eq:cjlm}), %
the kernel-based mirror design can incorporate nonlinear dependence effectively. 

\subsection{Decomposition of log-density function}
For notational simplicity, we write
$(U,V,W) = (X_j+cZ,X_j-cZ, X_{-j})$ and denote their  log-transferred joint density function as $\eta$, which belongs to a tensor product reproducing kernel Hilbert space (RKHS) $\cH$ \cite{lin2000tensor}.  We define $\cH=\cH^{U}\otimes\cH^{V}\otimes\cH^{W}$, where $\cH^{U}$, $\cH^{V}$ and $\cH^{W}$ are marginal RKHSs and ``$\otimes$" denotes the tensor product of two vector spaces.
We decompose $\eta$ as
\begin{multline}\label{eq:deceta}
\eta(u,v,w) =\eta_{U}(u) + \eta_{V}(v) + \eta_{W}(w) +\eta_{U,W}(u,w) \\
+ \eta_{V,W}(v,w) +  \eta_{U,V}(u,v)+ \eta_{U,V,W}(u,v,w).
\end{multline}
The uniqueness of the decomposition is guaranteed by the probabilistic decomposition of $\cH$. 

For simplicity, we  use the Euclidean space as an example to illustrate the basic idea of tensor sum decomposition, which is also known as ANOVA decomposition in linear models. For example, for the $d$-dimensional Euclidean space, we let $f$ be a vector and let $f(x)$ be its $x$-th entry,  $x=1,\dots, d$. Suppose $\cA$ is the  average operator defined as $\cA f(x) = \inner{\delta}{f}$, where $\delta=(\delta_1,\dots, \delta_d)$ is a probability vector (i.e., $\delta_i\geq 0$, and  $\sum \delta_i =1$).
The tensor sum decomposition of the Euclidean space $\bbR^d$ is
\begin{equation*}
\bbR^d = \bbR^d_0 \oplus \bbR^d_1 :=\{\delta\}\oplus\{f\in\bbR^d\, |\, \sum_{x=1}^d \delta_x f(x) =0 \},
\end{equation*}
where the first space is called the grand mean and the second space is called the main effect. Then, we construct the kernel for $\bbR^d_0$ and $\bbR^d_1$ as in  Lemma A.2 in the Supplementary Material (SM).

However, in RKHS of infinite dimension, the grand mean is not a single vector. Here, we set the average operator $\cA_l$ as $\cA_l := f \to E_x f(x) = E_x \inner{K^l_x}{f}_{\cH^l} = \inner{E_x K^l_x}{f}_{\cH^l}$ where $K^l$ is the kernel function in $\cH^l$ and the first equality is due to the reproducing property.  $E_x K^l_x$ plays the same role as $\delta$  in Euclidean space. Then we have the tensor sum decomposition of marginal RKHS $\cH^{l}$ defined as  
\begin{equation}\label{eq:dec}
\cH^{l} = \cH^{l}_0 \oplus \cH^{l}_1 := \{E_x K_x\} \oplus \{f\in \cH^{l} : \cA_l f=0\}.
\end{equation}
Following the same fashion, we call $\cH^l_0$ as the grand mean space and $\cH^l_1$ as the main effect space. Note that $E_x K^l_x$ is also known as the kernel mean embedding, which is well established in the statistics literature \cite{berlinet2011reproducing}.   We show the kernel functions for $\cH^l_0$ and $\cH^l_1$ in Lemma A.3 (see SM for details).

Following \cite{gu2013smoothing}, we apply the distributive law and have the decomposition of $\mathcal{H}$ as
\begin{align}
\mathcal{H} =& (\cH^{\cX}_0\oplus\cH^{\cX}_1)\otimes(\cH^{\cY}_0\oplus\cH^{\cY}_1)\otimes(\cH^{\cZ}_0\oplus\cH^{\cZ}_1)\nonumber\\
\equiv & \mathcal{H}_{000}\oplus\mathcal{H}_{100}\oplus\mathcal{H}_{010}\oplus\cH_{001}\oplus\cH_{110}\oplus\cH_{101} \nonumber\\ 
&\oplus\cH_{011}\oplus\cH_{111}\label{eq:anova}
\end{align}
where $\cH_{ijk} = \cH^{U}_i\otimes\cH^{V}_j\otimes \cH^{W}_k$. We show the kernel functions for each subspace in Lemma A.4 (see SM).  Each component in (\ref{eq:deceta}) is the projection of $\eta$ on the corresponding subspace in (\ref{eq:anova}).
Thus, the decomposition of the log-density function in (\ref{eq:deceta}) is unique.

We introduce the following lemma to establish a sufficient and necessary condition for  the conditional independence of $U$ and $V$ given $W$ based on the decomposition in (\ref{eq:deceta}).
\begin{lem}\label{lem:1}
	Assume that the  log-joint-density function $\eta$ of $(U,V,W)$ belongs to a tensor product RKHS. $U$ and $V$ are conditional independent if and only if $\eta_{U,V}+ \eta_{U,V,W} =0$.
\end{lem}
Here $\eta_{\cC}$ is the function only of variables in the set $\cC$. The proof of this lemma is given in SM.
Lemma \ref{lem:1} implies that the following two hypothesis testing problems are equivalent:
\begin{equation}\label{eq:hy1}
H_0: U\ci V \mid W \mbox{ vs. } H_1: U\nci V\mid W
\end{equation}
and
\begin{equation}\label{eq:hy2}
H_0: \eta\in \cH_0 \mbox{ vs. } H_1: \eta\in \cH / \cH_0,
\end{equation}
where $\cH_0: =\cH_{000}\oplus\cH_{100}\oplus\cH_{010}\oplus\cH_{001}\oplus\cH_{011}\oplus\cH_{101}$. Compared to (\ref{eq:hy1}),  a key advantage of  (\ref{eq:hy2}) is that we are able to specify the function space of the log-transferred density under both null and alternative.

\subsection{Kernel-based Conditional Dependence Measure}
Let %
$\bt_i = (U_i, V_i , W_i)$, $i=1,\ldots,n$, be \textit{i.i.d.} observations generated from the distribution of $\cT = (\cX,\cY,\cZ)$. The log-likelihood-ratio functional is
\begin{align}
\begin{split}\label{eq:lr}
LR_n(\eta) &= \ell_{n}(\eta) - \ell_{n}(P_{\cH_0}\eta) \\
&= -\frac{1}{n}\sum_{i=1}^{n}\{\eta(\bt_i) - P_{\cH_0}\eta(\bt_i)\}, \,\,  \eta\in\cH,
\end{split}
\end{align}
where $P_{\cH_0}$ is a projection operator from $\cH$ to $\cH_0$. Using the reproducing property, we rewrite (\ref{eq:lr}) as
\begin{equation}
LR_n(\eta) = -\frac{1}{n}\sum_{i=1}^n \left\{ \inner{K^{\cH}_{\bt_i}}{\eta}_{\cH} - \inner{K^{\cH_0}_{\bt_i}}{\eta}_{\cH} \right\},
\end{equation} 
where $K^{\cH}$ is the kernel for $\cH$ and $K^{\cH_0}$ is the kernel for $\cH_0$. Then, we calculate the Fr\'{e}chet derivative of the likelihood ratio functional as
\begin{align}
D LR_{n}(\eta) \Delta \eta &= \inner{\frac{1}{n}\sum_{i=1}^n (K^{\cH}_{\bt_i} - K^{\cH_0}_{\bt_i})}{\Delta\eta}_{\cH} \nonumber\\
&=  \inner{\frac{1}{n}K^{1}_{\bt_i}}{\Delta\eta}_{\cH}
\end{align}
where $K^{1}$ is the kernel for $\cH_{110} \oplus\cH_{111}$. The score function $\frac{1}{n}K_{\bt_i}^1$ is the first order approximation of the likelihood ratio functional.
We define our kernel-based conditional measure as the squared norm of the score function of the likelihood ratio functional as
\begin{equation}\label{eq:score}
[I_j^K(c)]^2 = \left\|\frac{1}{n}\sum_{i=1}^n K^1_{\bt_i} \right\|_{\cH}^2,
\end{equation}
which, by the reproducing property,  can be expanded as
\begin{equation}\label{eq:sn}
[I_j^K(c)]^2 = \frac{1}{n^2}\sum_{i=1}^n\sum_{j=1}^n K^1(\bt_i, \bt_j).
\end{equation}
The construction of $[I_j^K(c)]^2$ is related to the kernel conditional independence test \cite{fukumizu2008kernel}, which generalizes the conditional covariance matrix to a conditional covariance operator in RKHS. However, the calculation of the norm of the conditional covariance operator involves the inverse of an $n\times n$ matrix, which is expensive to obtain when the sample size $n$ is large. But we will show that our proposed score test statistics only involve  matrix multiplications.

We introduce a matrix form of the squared norm of score function to facilitate the computation process. In (\ref{eq:sn}), $[I_j^N(c)]^2$ is determined by the kernel on $\cH^{U}_1 \otimes \cH^{V}_1 \otimes \cH^{W}_0 \oplus \cH^{U}_1 \otimes \cH^{V}_1 \otimes \cH^{W}_1$. Thus, by Lemma A.2 and Lemma A.3, we can rewrite (\ref{eq:sn}) as
\begin{align}\label{eq:snmatrix}
[I_j^K(c)]^2 = \frac{1}{n^2} [(HK^{U}H) \circ (HK^{V}H) \circ K^{W} ]_{++}
\end{align}
where $H= I_n -  \frac{1}{n} \mathbf{1}\mathbf{1}^T$, $I_n$ is a $n\times n$ identity matrix and $\mathbf{1}_n$ is a $n\times 1$ vector of ones, and $[A]_{++}=\sum_{i=1}^n\sum_{j=1}^n A_{ij}$.The most popular  kernel choices are Gaussian  and polynomial kernels. With parallel computing, the computational complexity for using either is approximately linear in $n$.

\begin{figure}[!t]
	\centering
	\includegraphics[width=0.45\textwidth]{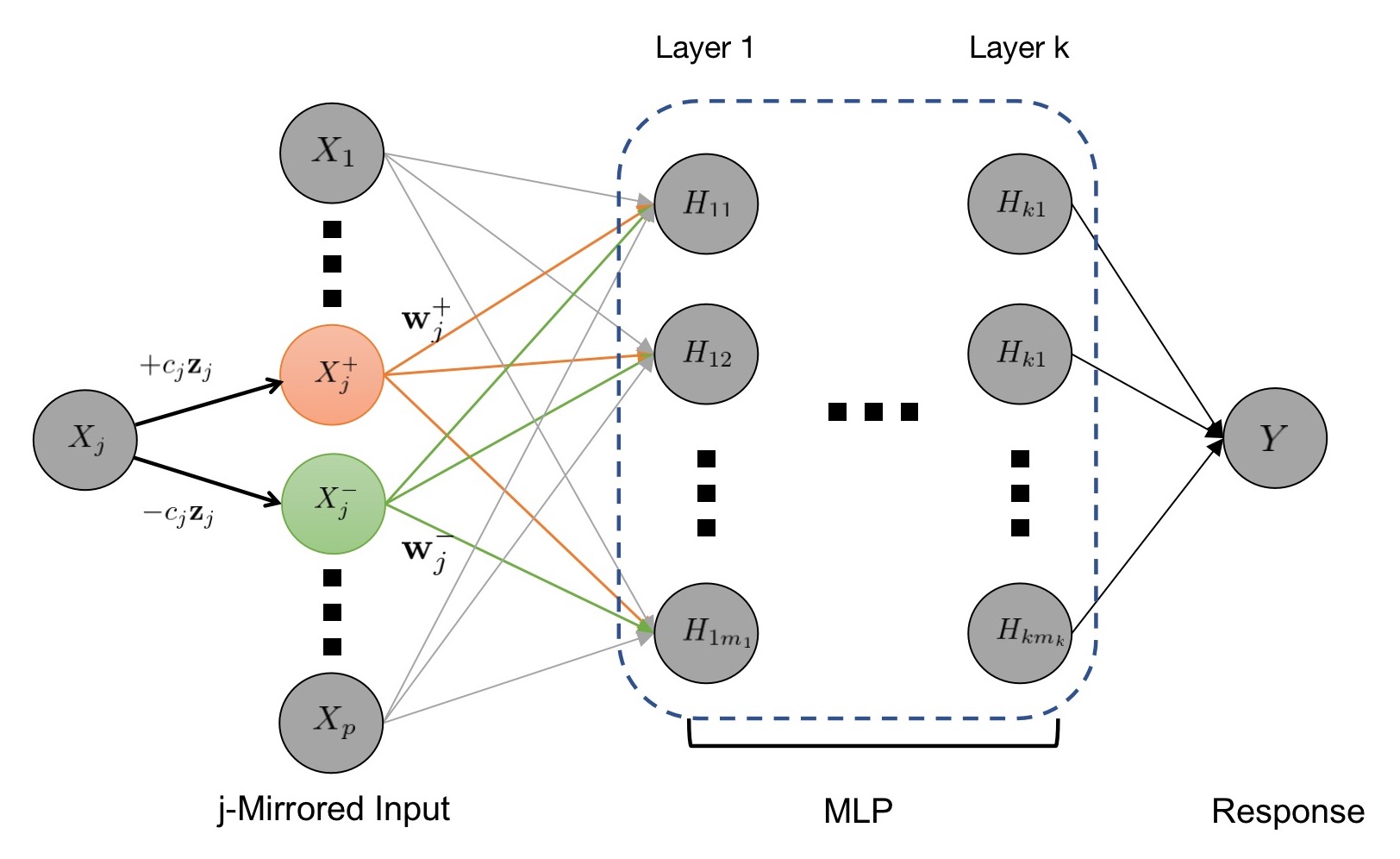}
	\caption{Indivdual neural Gaussian mirror in DNN}
	\label{fig:iNGM}
\end{figure}

\begin{figure}[!t]
	\centering
	\includegraphics[width=0.45 \textwidth]{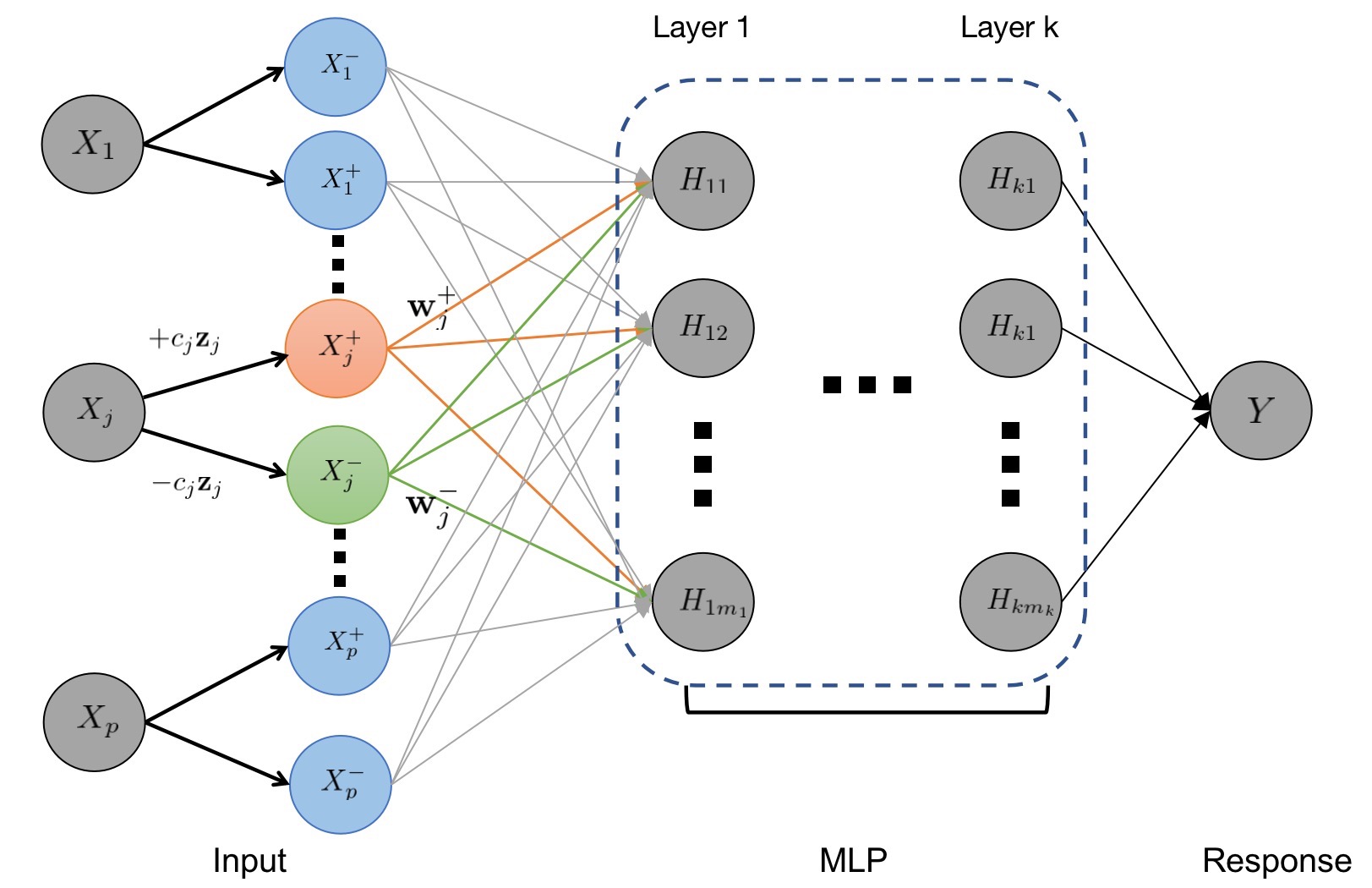}
	\caption{Simultaneous neural Gaussian mirror in DNN}
	\label{fig:sNGM}
\end{figure}

\subsection{Mirror Statistics in DNNs}

In order to construct mirror statistics in DNNs,  we introduce an importance measure for each input feature as a generalization of the  connection weights method proposed by \cite{feature}.
We consider a fully connected  multi-layer perceptron 
(MLP) with $k$ hidden layers denoted as $(H^{(t)}_{1},\dots, H^{(t)}_{n_t})$ for $t=1,\dots,k$.
The connection between the $t$-th and $(t+1)$-th layers can be expressed as 
\begin{equation*}
H_{j}^{(t+1)} = g_{t+1}(\sum_{i=1}^{n_t}w_{ij}^{(t)} H_{i}^{(t)}), \ \mbox{for } \ j= 1, \dots, n_{t+1},
\end{equation*}
where $g_{t+1}$ is the activation function. We let $t=0$ and $t=k+1$ denote the input and output layers, respectively.

For a path 
$\bm{l}=(X_j,H_{i_1}^{(1)},...,H_{i_k}^{(k)},y)$ through the network, We define its accumulated weight as
\begin{align*}
\delta(\bm{l})= \omega_{ji_1}^{(0)}\omega^{(k)}_{i_k1}\prod_{t=1}^{k-1} \omega^{(t)}_{i_ti_{t+1}}.
\end{align*} 
Let $\bm{\Omega}^{(0)}_j=(\omega_{j1}^{(0)},\dots,\omega_{jn_1}^{(0)})$ be the weight vector connecting $X_j$ with the first layer. We define the feature importance of $X_j$ as
\begin{equation}\label{eq:importance}
L(X_j) = \sum_{\bm{l}\in P_j} \delta(\bm{l}) = \langle \bC, \bm{\Omega}^{(0)}_j \rangle, 
\end{equation}
where $P_j$ is the set consisting of all paths connecting $X_j$ to $y$ with one node in each layer and $\bC = \prod_{t=1}^{k} \bm{\Omega}^{(t)}$, where $\bm{\Omega}^{(t)}_{ij} = \omega_{ij}^{(t)}\in \RR^{n_t \times n_{t+1}}$ for $i=1,\dots,n_t$ and $j=1,\dots,n_{t+1}$. The gradient w.r.t. each input feature $\partial y/\partial X_j$ in \cite{hechtlinger2016interpretation} is a modified version of $L(X_j)$ (see the SM for details).

Based on (\ref{eq:importance}), we define mirror statistics as
\begin{align}\label{eq:mstat}
M_j = |L(X_j^{+})+L(X_j^{-})| - |L(X_j^{+})-L(X_j^{-})|,
\end{align} 
and use (\ref{eq:fdphat}) to estimate FDP. 
By setting a predefined error rate $q$, a data adaptive threshold is given by
$\tau_q = \min_{t}\{t>0:\widehat{FDP}(t) \leq q\}$
and $\hat{S}_1 = \{j:M_j \geq \tau_q\}$ is the set of selected features.

\section{Implementations}
\subsection{Two forms of neural Gaussian mirror}
The NGM can be realized in two ways: individual or simultaneous. As shown in Figure~\ref{fig:iNGM},  the {\it individual neural Gaussian mirror} (INGM) is constructed by treating one feature a time. More specifically, 
for feature $\bX_j$, we create $\bX_j^{+}$ and $\bX_j^{-}$  with $c_j$ minimizing $[I_j^K(c)]^2$. 
Then, we set the input layer as $(X_1,\dots, X_j^+, X_j^-,\dots, X_p)$  and fit the MLP as shown in Figure \ref{fig:iNGM}.  For any predefined error rate $q$, we select feature by Algorithm \ref{alg:1}.

\begin{algorithm}[ht]
	\caption{Individual NGM: feature selection in deep neural networks}\label{alg:1}
	\textbf{Input:} Fixed FDR level $q$, $(\bx_i,y_i)$, $i=1,...,n$ with $\bx_i \in \mathbb{R}^{p}$, $y_i \in \mathbb{R}^{1}$\\
	\textbf{Output:} $\hat{S}_1$
	\begin{algorithmic}[1]
		\For{$i=1$ to $p$}
		\State Generate $\bZ_j \sim \mathcal{N}(0,\bI_n)$
		\State Compute $c_j = {\rm arg}\min_{c} [I_j^K(c)]^2$
		\State Create mirrored pair$ (\bX_j^{+},\bX_j^{-})=(\bX_j+c_j \bZ_j,\bX_j-c_j \bZ_j)$
		\State Train  MLP with $\bX^{(j)}=(\bX_j^{+},\bX_j^{-},\bX_{-j})$ as the input layer
		\State Compute mirror statistics $M_j$ by (\ref{eq:mstat})
		\EndFor
		\State Calculate $T= \min\{t>0: \widehat{FDP}(t) \leq q\}$.
		\State Return $\hat{S}_1 \leftarrow \{j:\;M_j \geq T\}$.
	\end{algorithmic}
\end{algorithm}

To increase the efficiency, we construct the {\it simultaneous neural Gaussian mirror } (SNGM),  by mirroring all the $p$ features at the same time as shown in Figure \ref{fig:sNGM}.  The input layer include all mirrored pairs, i.e., $(X_1^+, X_1^-,\dots, X_p^+, X_p^-) $.
The detailed algorithm is given in Algorithm \ref{alg:2}.

\begin{algorithm}[ht]
	\caption{SNGM: Feature selection with simultaneous-mirrored Pairs}\label{alg:2}
	\textbf{Input:} Fixed FDR level $q$, $(\bx_i,y_i)$, $i=1,...,n$ with $\bx_i \in \mathbb{R}^{p}$, $y_i \in \mathbb{R}^{1}$\\
	\textbf{Output:} $\hat{S}_1$
	\begin{algorithmic}[1]
		\For{$i=1$ to $p$}
		\State Generate $\bZ_j \sim \mathcal{N}(0,\bI_n)$
		\State Compute $c_j = {\rm arg}\min_{c} [I_j^K(c)]^2$
		\State Create mirrored pair $(\bX_j^{+},\bX_j^{-})=(\bX_j+c_j \bZ_j,\bX_j-c_j \bZ_j)$
		\EndFor
		\State Train MLP with input layer as $(\bX_1^+,\bX_1^-,\dots, \bX_p^+, \bX_p^-)$.
		\State Calculate $M_j$ by (\ref{eq:mstat}) for $j=1,\dots,p$.
		\State Calculate $T= \min\{t>0: \widehat{FDP}(t) \leq q\}$.
		\State {\bfseries Output} $\hat{S}_1 \leftarrow \{j:\;M_j \geq T\}$.
	\end{algorithmic}
\end{algorithm}

\subsection{Screening}
To further reduce the computational cost,  we propose a screening step based on the rank of feature importance measure defined in (\ref{eq:importance}). The screening procedure is inspired by the RANK method proposed by \cite{fan2019rank, lu2018deeppink}, which uses part of the data for estimating the precision matrix and subset selection and leaves the remaining data for controlled feature selection.

In the screening procedure, we first randomly select $[n/3]$ samples $(\bX^{(1)},\by^{(1)})$ to train a neural network.  We calculate feature importance measure $\{L(\bX^{(1)}_j)$ for $ j=1,2,\dots,p\}$ and rank  their absolute values from large to small as $|L|_{(1)} \geq |L|_{(2)} \geq \dots \geq |L|_{(p)}$. We denote $\hat{S}_m = \{j:L(\bX^{(1)}_j) \geq |L|_{(m)}\}$ as our screened set. Empirically, we can set $m$ as $[n/2]$ or $[2n/log(n)]$. The screening can be implemented before the INGM and the SNGM algorithms to save computational costs. we call the INGM with screening and SNGM with screening as S-INGM and S-SNGM, respectively.

\begin{figure*}[!t]
	\centering
	\includegraphics[width=0.95\textwidth]{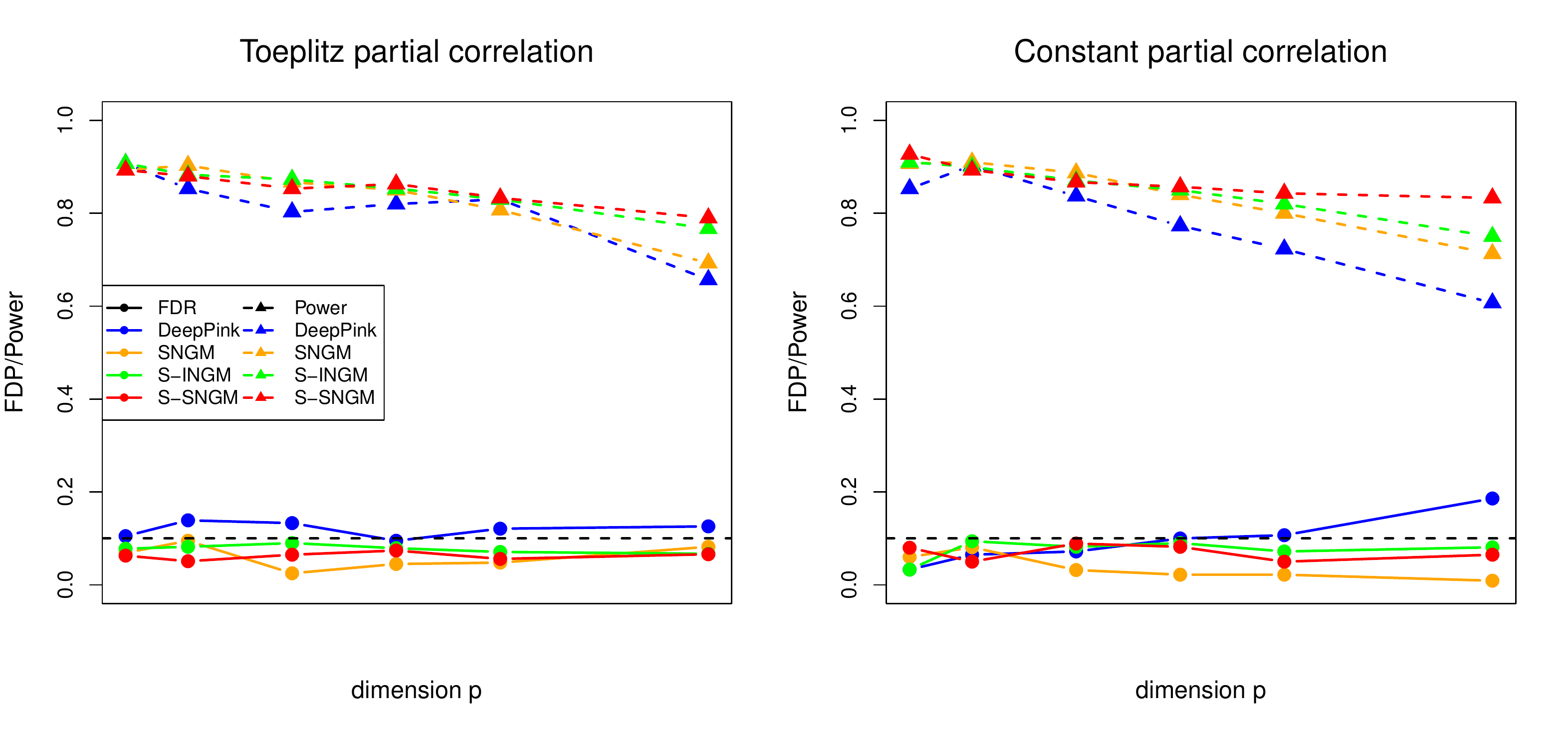}
	\caption{Performance comparison of NGMs and DeepPINK with predefined FDR level $q=0.1$ in linear models with two partial correlation structures}
	\label{fig:plot_linear}
\end{figure*} 
\begin{table*}[!t]
	\centering
	\fontsize{8}{8}\selectfont
	\begin{threeparttable}
		\caption{Varying single-index link functions with two PC (partial correlation) structures (q=0.1)}
		\label{tab:si}
		\begin{tabular}{ccccccccccc}
			\toprule
			\multirow{2}{*}{Setting} 
			&\multirow{2}{*}{$n=1000$} & \multirow{2}{*}{Link function} & \multicolumn{2}{c}{S-SNGM} & \multicolumn{2}{c}{SNGM}& \multicolumn{2}{c}{S-INGM}& \multicolumn{2}{c}{DeepPink}\\
			\cmidrule(lr){4-5} \cmidrule(lr){6-7} \cmidrule(lr){8-9} \cmidrule(lr){10-11}
			& & &FDR &Power &FDR &Power &FDR &Power &FDR &Power\\
			\midrule
			\multirow{6}{*}{Toeplitz PC}&\multirow{3}{*}{$p=500$} & $f_1(t)=t+sin(t)$ &0.071 &0.867 &0.045 &0.857 &0.057 &0.900 &0.085 &0.900 \\
			&& $f_2(t)=0.5t^3$ &0.092 &0.873 &0.080 &0.843 &0.061 &0.830 &0.155 &0.413 \\
			&& $f_3(t)=0.1t^5$ &0.059 &0.865 &0.071 &0.847 &0.132 &0.807 &0.148 &0.333 \\
			\cmidrule{2-11}
			&\multirow{3}{*}{$p=2000$} & $f_1(t)=t+sin(t)$ &0.085 &0.843 &0.081 &0.820 &0.084 &0.837 &0.126 &0.817 \\
			&& $f_2(t)=0.5t^3$ &0.076 &0.788 &0.082 &0.587 &0.081 &0.810 &0.149 &0.320 \\
			&& $f_3(t)=0.1t^5$ &0.093 &0.737 &0.108 &0.530 &0.185 &0.620 &0.477 &0.053 \\
			\midrule
			\multirow{6}{*}{Constant PC}&\multirow{3}{*}{$p=500$} & $f_1(t)=t+sin(t)$ &0.057 &0.880 &0.067 &0.877 &0.070 &0.900 &0.136 &0.86 \\
			&& $f_2(t)=0.5t^3$ &0.084 &0.870 &0.047 &0.867 &0.057 &0.847 &0.084 &0.587 \\
			&& $f_3(t)=0.1t^5$ &0.066 &0.867 &0.066 &0.847 &0.027 &0.802 &0.143 &0.46 \\
			\cmidrule{2-11}
			&\multirow{3}{*}{$p=2000$} & $f_1(t)=t+sin(t)$ &0.055 &0.857 &0.012 &0.833 &0.057 &0.813 &0.094 &0.800 \\
			&& $f_2(t)=0.5t^3$ &0.064 &0.787 &0.016 &0.613 &0.104 &0.803 &0.174 &0.313 \\
			&& $f_3(t)=0.1t^5$ &0.080 &0.743 &0.020 &0.548 &0.111 &0.607 &0.412 &0.147 \\
			\bottomrule
		\end{tabular}
	\end{threeparttable}
\end{table*}

\section{Numerical simulations}
\begin{figure*}[!t]
	\centering
	\includegraphics[width=0.95\textwidth]{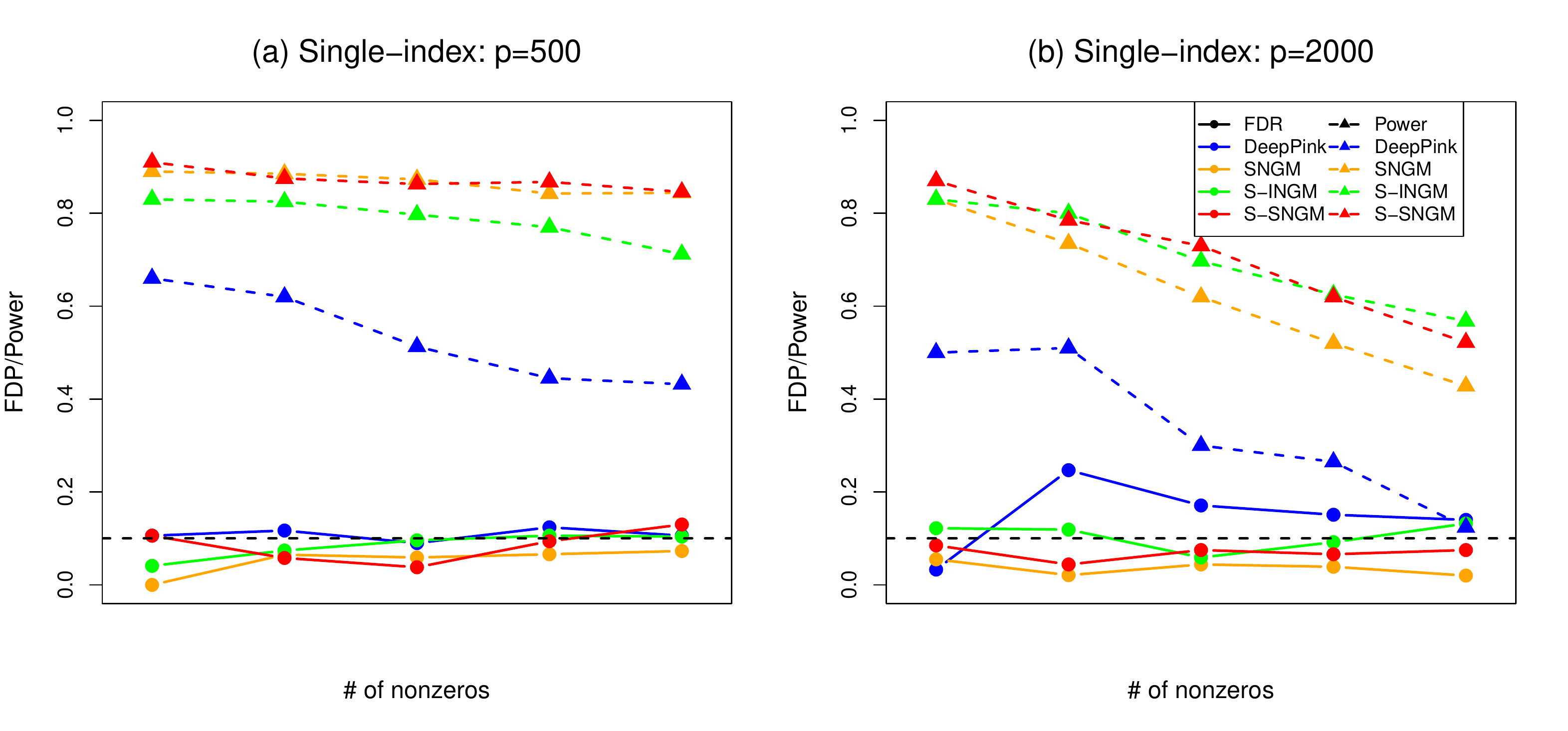}
	\caption{Performance comparison of NGM and DeepPINK with predefined FDR level $q=0.1$ under single-index model with link function $f_2$. The triangle dots denote power and round dots denote FDR. (a). p=500; (b). p=2000}
	\label{fig:plot_si}
\end{figure*}  
\begin{figure}[!t]
	\centering
	\includegraphics[width=0.5\textwidth]{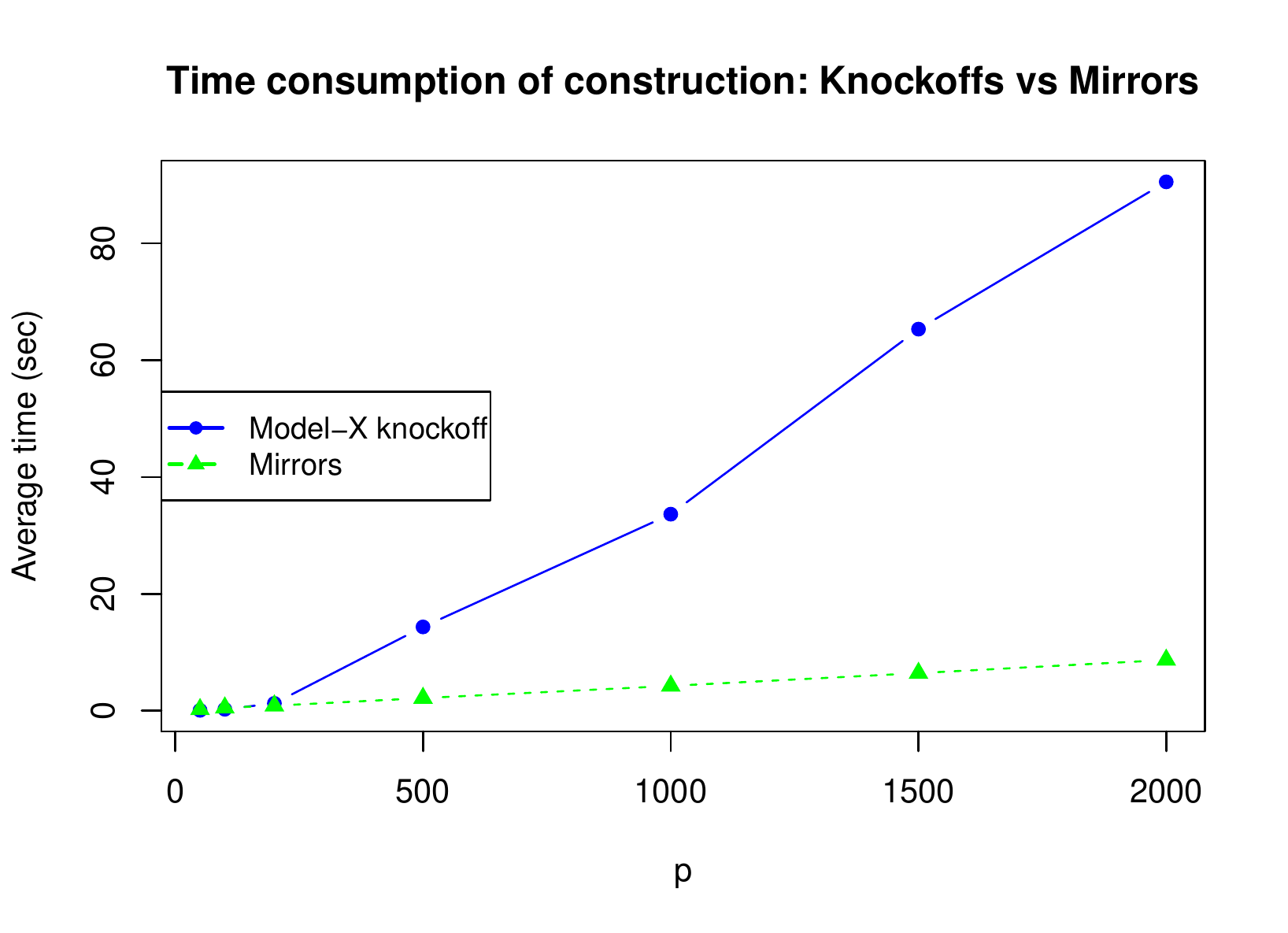}
	\caption{Computational time for constructing input variables: Mirrors vs Model-X Knockoffs}
	\label{fig:time}
\end{figure}
We choose an MLP structure with two hidden layers, with $N_1=20log(p)$ and $N_2=10log(p)$ hidden nodes respectively.%
We compare the selection power and FDR of NGMs with the DeepPINK, and conduct experiments in both simulated and real-data settings with $20$ repetitions. 
In the simulation studies, we consider data from both linear models and nonlinear models. In each setting we consider two covariance structures for $\bX$: the Toeplitz partial correlation structure and the constant partial correlation structure. The Toeplitz partial correlation structure has its  precision matrix (i.e., inverse) as $\bOmega= (\rho^{|i-j|})$. The constant partial correlation structure has its precision matrix as $\bOmega=(1-\rho)\bI_p+\rho \bJ_p$ where $\bJ_p = \frac{1}{p}\bm{1}_p\bm{1}_{p}^T$, where $\rho=0.5$.

\subsection{Linear models}
First, we examine the performance  of all methods in linear models: $y_i=\bbeta^\top \bx_i+\epsilon_i,  \ \epsilon_i\stackrel{\mbox{\small{i.i.d.}}}{\sim} N(0,1)$,
for $i=1,\dots, n$. We randomly set $k=30$ elements in $\bbeta$ to be nonzero and generated from $\mathcal{N}(0,(20\sqrt{log(p)/n})^2)$ to mimic various signal strengths in real applications.

As shown in Figure \ref{fig:plot_linear}, NGMs control the FDR at $q=0.1$ and have a higher power than the DeepPINK. In the constant partial correlation setting with partial correlation $\rho=0.5$, DeepPINK shows a power loss when $p$ is larger than $1500$ since the minimum eigenvalue of the correlation matrix is $1/(1+(p-1)\rho)$, which approaches 0 as $p \rightarrow \infty$ and makes  important features highly correlated with their knockoff counterparts. By introducing a random perturbation, which reduces the correlation between mirror statistics, NGMs control FDR at the level of $q=0.1$ and maintain power around $0.8$ in both correlation structures. 

In high-dimensional cases when $p \geq 2000$, the NGMs, especially the SNGM and S-SNGM, are  computationally more efficient than DeepPINK. In addition, we note that the difference in time consumption between DeepPINK and SNGM mainly lies in the construction of input variables. %
As shown in \ref{fig:time}, computational time of constructing knockoff variables in DeepPINK increases rapidly when the number of features grows.%
In contrast, for SNGM and S-SNGM, the construction of mirrored input is parallel for each feature, which makes the computation time less affected by the increase of dimension $p$.

\subsection{Single-index models}
We further test the performance of the proposed  methods and DeepPINK in nonlinear cases such as single-index models.
In our experiment, we choose three nonlinear link functions: $f_1(t)=t+sin(t)$, $f_2(t)=0.5t^3$ and $f_3(t)=0.1t^5$. We try both high-dimensional cases with $p=2000$ and low-dimensional cases with $p=500$. As shown in Table \ref{tab:si}, NGMs and DeepPINK all have desirable performances similar to that in linear models since the first function  $f_1(t)=t+sin(t)$ is dominated by its linear term. However, in the latter two cases where $f$ is a polynomial, Table \ref{tab:si} shows that DeepPINK suffers a loss of power. In all the three  cases, NGMs are capable of maintaining the power at a higher level than DeepPINK.

To study the influence of sparsity, we further consider an experiment with constant partial correlation $\rho=0.5$ and varying the number of important features from $10$ to $50$ skipping by $10$. For varying sparsity levels, as shown in Table \ref{tab:si}, NGMs  maintain high power with controlled error rate in both the low dimensional case with $p=500$ and the high dimensional setting with $p=2000$, whereas DeepPINK tends to loss power when the sparsity level is high.

\subsection{Real data design: tomato dataset}
We consider a panel of 292 tomato accessions in \cite{bauchet2017use}. The panel includes breeding materials that are built and characterized with over 11,000 SNPs. Each SNP is coded as 0, 1 and 2 to denote the homozygous (major), heterozygous, and the other homozygous (minor) genotypes, respectively. To evaluate the performance of NGMs and DeepPINK under non-Gaussian designs, we randomly select $p$ ranging from 100 to 1000 SNPs  to form our design matrix, and generate the response $y_i$ in the same way as with the  simulated examples. This simulation is replicated $20$ times independently, to which each method in consideration is applied.

We set the pre-defined FDR rate at $q=0.2$ and compare the empirical FDR and power of NGMs with those of DeepPINK. As shown in table \ref{tab:1},  the S-INGM method shows the lowest FDR and highest power when $p=100$ and 500.  When $p=$1000 and 2000, S-INGM is more powerful than DeepPINK but the its FDR inflates slightly. S-SNGM  is  computationally more efficient, but with slightly inferior performances,  than S-INGM. 

\begin{figure}[!t]
	\centering
	\includegraphics[width=0.5\textwidth]{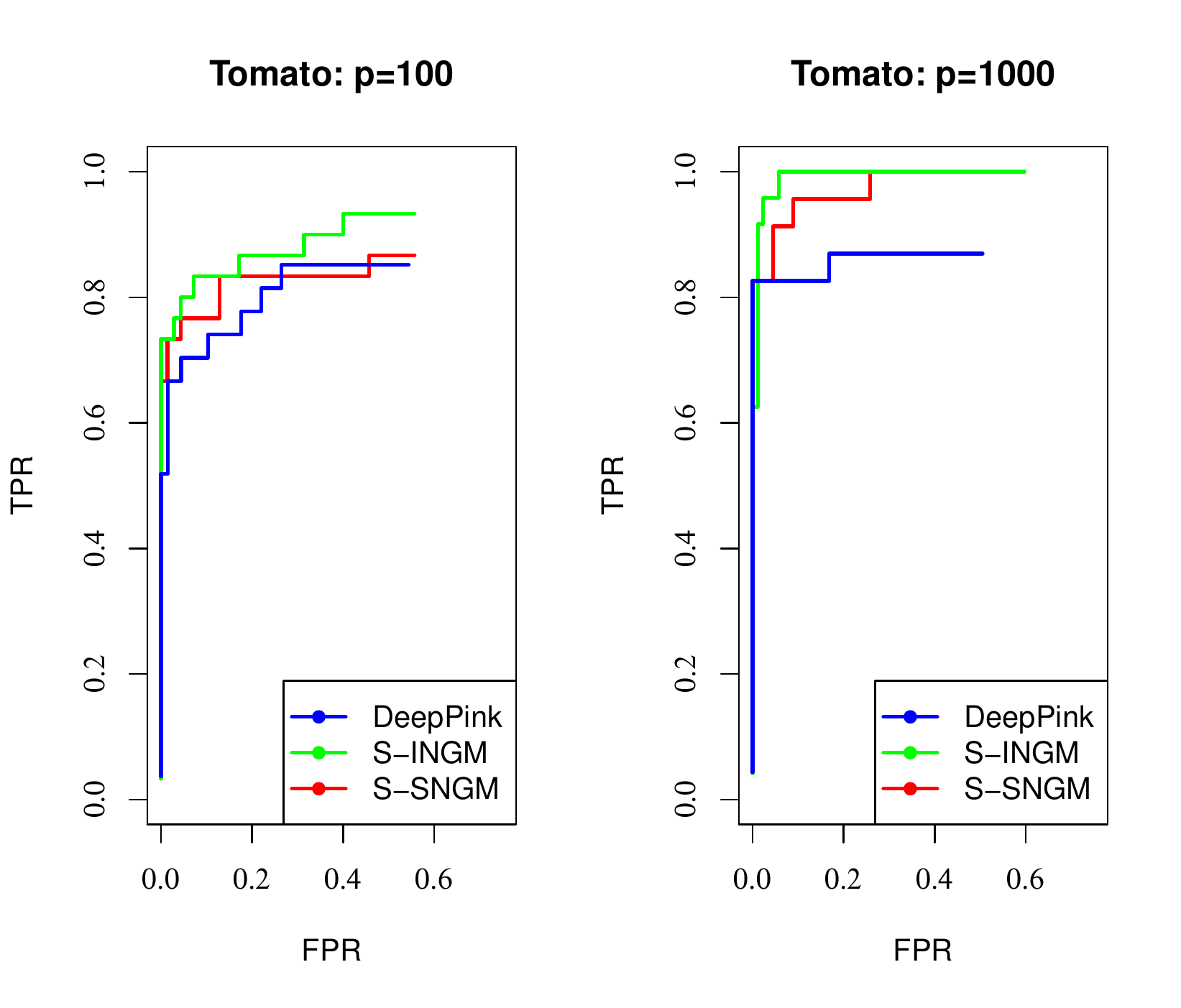}
	\caption{ROC curve for performance on tomato dataset}
	\label{fig:roc}
\end{figure}

To provide a better visualization of the trade-off between the error rate and power, we plot the ROC curve for the three methods. %
Specifically the ROC curve is plotted with true positive rate 
\begin{align}
    {\rm TPR}=\frac{{\rm TP}}{{\rm TP+FN}}
\end{align} against the false positive rate 
\begin{align}
    {\rm FPR}=\frac{{\rm FP}}{{\rm FP+TN}}.
\end{align}
A method with a larger area under the curve (AUC) is regarded as a method with better trade-off between power and FDR.
In both low-dimensional and high-dimensional settings, Figure \ref{fig:roc} shows that NGMs have a larger  AUC than DeepPINK. 

\begin{table}[!t]
	\centering
	\fontsize{8}{12}\selectfont
	\begin{threeparttable}
		\caption{Performance comparison with tomato design (q=0.2, k=30)}\label{tab:1}
		\label{tab:tomato}
		\begin{tabular}{lcccccc}
			\toprule
			\multirow{2}{*}{$n=292$} 
			& \multicolumn{2}{c}{S-INGM}& \multicolumn{2}{c}{S-SNGM}& \multicolumn{2}{c}{DeepPink}\\
			\cmidrule(lr){2-3} \cmidrule(lr){4-5} \cmidrule(lr){6-7}
			&FDR &Power &FDR &Power &FDR &Power \\
			\midrule
			$p=100$ &0.171 &0.893 &0.131 &0.690 &0.182 &0.607 \\
			$p=500$ &0.156 &0.964 &0.129 &0.586 &0.209 &0.403 \\
			$p=1000$ &0.229 &0.942 &0.281 &0.520 &0.145 &0.490 \\
			$p=2000$ &0.296 &0.864 &0.347 &0.577 &0.244 &0.453 \\
			\bottomrule
		\end{tabular}
	\end{threeparttable}
\end{table}

\section{Discussion}
In this paper, we propose NGMs for feature importance assessment and controlled selection in neural network models, which is an important aspect for model interpretation.  Even in situations with no need of  explicit feature selections, having a good rank of the predictors according to their influence on the output can be very helpful for practitioners to prioritize their follow-up work.
We emphasize that our method does not require any distributional assumption on $\bX$, thus is widely applicable to a broad class of neural network models including those with discrete or categorical features. In addition, the mirror design can be generated to convolutional neural networks (CNN) to measure the stability of the filters. For example, for each original CNN slice in the input tensor, we can create its corresponding mirrored slice by adding and subtracting a slice of random perturbations, and contrast the
influences of the original and mirrored slices.  detailed study in this direction is deferred to a future work. 
The data used in the real example is available in (\url{ftp.solgenomics.net/manuscripts/Bauchet\_2016/}).

\appendix
\section*{Supplementary materials}
\section{Test statistics for conditional independence}
To simplify the notations, let 
\[
(U,V,W) = (X_j^{+},X_j^{-}, X_{-j})
\]

\begin{lem}
	$U$ and $V$ are conditional independent if and only if $\eta_{U,V}(u,v)+ \eta_{U,V,W}(u,v,w) =0$
\end{lem}

\noindent\underline{\textit{Proof}}

Suppose $f$ is the joint density function for $(U,V,W)=(X_j^{+}, X_j^{-}, X_{-j})$, then it can be decomposed into $2^3=8$ factors: 
\begin{align}
f(u,v,w) &= \exp(\eta_{U}(u) + \eta_{V}(v) + \eta_{W}(w) \nonumber\\ & +\eta_{U,W}(u,w) + \eta_{V,W}(v,w)+  \eta_{U,V}(u,v) \nonumber\\ & + \eta_{U,V,W}(u,v,w)),
\end{align}
where $\eta$ represents log-transformed density and $\eta_{A}(\alpha)$ means the part as the function only of $\alpha$. Then
\begin{align}
\eta(u,v,w)  & =\eta_{U}(u) + \eta_{V}(v) + \eta_{W}(w) \nonumber\\ &+\eta_{U,W}(u,w) + \eta_{V,W}(v,w) +  \eta_{U,V}(u,v) \nonumber\\ &+ \eta_{U,V,W}(u,v,w)
\end{align}

To assess the dependence between $(U,V)$ conditioned on $ W$, we derive the conditional density as
\begin{align}
& f_{(U,V|W)}(u,v|w) = \frac{e^{\eta(u,v,w)}}{\int_\cU\int_\cV e^{\eta(u,v,w)}}\nonumber\\
& = C(w)\cdot e^{\eta_{U}(u)+\eta_{U,W}(u,w)} \cdot \nonumber\\ & ~~~e^{\eta_{V}(v)+\eta_{V,W}(v,w)} \cdot e^{\eta_{U,V}(u,v)+ \eta_{U,V,W}(u,v,w)}
\end{align}
where $C(w)$ denotes the denominator as the marginal density of $W$. Therefore $U \ci V|W$ if and only if any interaction of $U$ and $V$ is constants, i.e. $\eta_{U,V}+ \eta_{U,V,W}=0.$
$\hfill\blacksquare$ 

\section{Importance measurement in DNN}

In order to construct mirror statistics in DNNs,  we first adopt an importance measure for each input feature.

We consider a fully connected  multi-layer perceptron 
(MLP) with $k$ hidden layers denoted as $(H^{(t)}_{1},\dots, H^{(t)}_{n_t})$ for $t=1,\dots,k$.
The output of the ($t+1$)-th layer can be described as
\begin{equation*}
H_{j}^{(t+1)} = g_{t+1}(\sum_{i=1}^{n_t}w_{ij}^{(t)} H_{i}^{(t)}), \ \mbox{for } \ j= 1, \dots, n_{t+1},
\end{equation*}
where $g_{t+1}$ is the activation function. Note that $t=0$ denotes the input layer and $t=k+1$ denotes the output layer.

We define 
\begin{align*}
\delta(\bl(X_j,H_{i_1}^{(1)},...,H_{i_k}^{(k)},y)) = \omega_{ji_1}^{(0)}\omega^{(k)}_{i_k1}\prod_{t=1}^{k-1} \omega^{(t)}_{i_ti_{t+1}}
\end{align*} as the accumulated weight of the path $\bl$.

Let $\bOmega^{(0)}_j=(\omega_{j1}^{(0)},\dots,\omega_{jn_1}^{(0)})$ as the weight vector connecting $X_j$ with the first layer. We define the feature importance of $j$th feature as
\begin{align}\label{eq:import}
L(X_j) = \sum_{\bl\in P_j} \delta(\bl) = \langle \bC, \bOmega^{(0)}_j \rangle, 
\end{align}
where $P_j$ is the set consisting of all paths connecting $X_j$ to $y$ with one node in each layer and $\bC = \prod_{t=1}^{k} \bOmega^{(t)}$, where $\bOmega^{(t)}_{ij} = \omega_{ij}^{(t)}\in \RR^{n_t \times n_{t+1}}$ for $i=1,\dots,n_t$ and $j=1,\dots,n_{t+1}$.

For $t=1,\dots,k$, define diagonal matrices 
\begin{align}
\bG^{(t)} = \diag ( & \frac{\partial g_{t}}{\partial h}(\sum_{i=1}^{n_{t-1}}w_{i1}^{(t-1)} H_{i}^{(t-1)}),\dots, 
\nonumber\\ & \frac{\partial g_{t}}{\partial h}(\sum_{i=1}^{n_{t-1}}w_{in_{t}}^{(t-1)} H_{i}^{(t-1)})).
\end{align}
Then, the gradient w.r.t. $X_j$ can be written as
\begin{align}
\frac{\partial y}{\partial X_j} = \bOmega^{(0)}_j\prod_{t=1}^{k} (\bG^{(t)}\bOmega^{(t)}),
\end{align}
which is a weighted version of $L(X_j)$.

\section{Perturbation with linear kernel}
Suppose $n$ observations $(\bx_i,y_i),i=1,2, \dots ,n$ are drawn from the model 
\[
y = \phi(\bx)^\top \bbeta +\epsilon
\]
where $\phi(\cdot)$ is a feature map and the kernel function $k(\cdot,\cdot)$ is define as 
\[
k(\bx,\bz) = \langle \phi(\bx), \phi(\bz) \rangle
\]
In the classical linear model $y = \bx^\top \bbeta +\epsilon$, $\phi$ is chosen to be the identity with a constant interception  and thus $k(\cdot,\cdot) = \langle \cdot, \cdot \rangle+{\rm const}$.

To specify the test statistics when adopting linear kernels
\begin{align}
&[I_j^K(c)]^2 \nonumber\\& = \frac{1}{n^2} [(\bH \bK^U \bH) \circ (\bH \bK^V \bH) \circ \bK^W]_{++}\nonumber\\
& = \frac{1}{n^2} \{[(\bH U)(\bH U)^\top] \circ [(\bH V)(\bH V)^\top] \nonumber\\ & ~~~~~~\circ (WW^\top)\}_{++}
\end{align}
where $(U,V,W)=(X_j^{+},X_j^{-},X_{-j})$ and $\bH= I_n -  \frac{1}{n} \mathbf{1}_n\mathbf{1}_n^\top$. To simplify notations, denote $U=\bH U$ and $V=\bH V$, with which
\begin{align}
& G(c) \triangleq n^2 [I_j^K(c)]^2\nonumber\\  &= \{[(X_j+cZ_j) (X_j+cZ_j)^\top] \nonumber\\ &~~~~~~\circ [(X_j-cZ_j) (X_j-cZ_j)^\top] \circ (WW^\top)\}_{++}. 
\end{align}
We take gradient of $G(c)$ with respect to $c$ and due to the community of operators:
\begin{align}
&\nabla_c G(c)\nonumber\\  &= \{\frac{\partial}{\partial c} ([(X_j+cZ_j) (X_j+cZ_j)^\top] \nonumber\\ & ~~~~~~\circ [(X_j-cZ_j) (X_j-cZ_j)^\top]) \circ (WW^\top) \}_{++}\nonumber\\
&= \{ \nabla_c [(X_j+cZ_j) (X_j+cZ_j)^\top] \nonumber\\ &~~~~~~\circ [(X_j-cZ_j) (X_j-cZ_j)^\top] \circ (WW^\top) \nonumber\\
&~~~+ [(X_j+cZ_j) (X_j+cZ_j)^\top] \nonumber\\ & ~~~~~~\circ \nabla_c [(X_j-cZ_j) (X_j-cZ_j)^\top] \circ (WW^\top)\}_{++}\nonumber\\
&= c\{[2c^2(Z_j Z_j^\top) \circ (Z_j Z_j^\top)+2(X_j X_j^\top) \circ (Z_j Z_j^\top) \nonumber\\
&~~~- (X_j Z_j^\top) \circ (X_j Z_j^\top) - (Z_j X_j^\top) \circ (Z_j X_j^\top) \nonumber\\ &~~~~~~-2(X_j Z_j^\top) \circ (Z_j X_j^\top)] \circ (WW^\top)\}_{++}.
\end{align}
Therefore, $\nabla_s G(s)=0$ leads to the solution as follow:
\begin{align}\label{kernelc}
c^{*}_j %
= \sqrt{\frac{\{A(X_j,Z_j) \circ (WW^\top)\}_{++}}{2[(Z_j Z_j^\top) \circ (Z_j Z_j^\top) \circ (WW^\top)]_{++}}},
\end{align}
with $A(X_j,Z_j) = (Z_j X_j^\top) \circ (Z_j X_j^\top) + (X_j Z_j^\top) \circ (X_j Z_j^\top)$.
And $c_j^{*}$ is the minimum point by computing the second-order derivatives w.r.t. $c$.

For the space $span\{W\}$, let $P_{W}$ be the projection operator, then $c_j$ in the Gaussian Mirror can be written as 
\begin{align}\label{gmc}
c_j^{\rm GM} = \sqrt{\frac{\Vert (I-P_{W})X_j \Vert^2}{\Vert (I-P_{W})Z_j \Vert^2}}.
\end{align}

We should note that, without the term $WW^\top$, $[(Z_j Z_j^\top) \circ (Z_j Z_j^\top)]_{++} = \Vert Z_j \Vert^4$, $[(Z_j X_j^\top) \circ (Z_j X_j^\top)]_{++} = [(X_j Z_j^\top) \circ (X_j Z_j^\top)]_{++} = \Vert X_j \Vert^2 \Vert Z_j \Vert^2$.
Therefore, comparing the form of \ref{kernelc} and \ref{gmc}, $c_j^{*}$ and $c_j^{\rm GM}$ are of the same scale which is validated in the simulation and they are equivalent with orthogonal designs.

\begin{table*}[!t]
	\centering
	\fontsize{8}{12}\selectfont
	\begin{threeparttable}
		\caption{Linear Model with varying constant partial correlation(q=0.1)}
		\label{tab:varying_cpc}
		\begin{tabular}{cccccccccc}
			\toprule
			\multirow{2}{*}{$n=1000$} & \multirow{2}{*}{partial correlation}& \multicolumn{2}{c}{S-SNGM} & \multicolumn{2}{c}{SNGM} & \multicolumn{2}{c}{S-INGM}& \multicolumn{2}{c}{DeepPink} \\

			\cmidrule(lr){3-4} \cmidrule(lr){5-6} \cmidrule(lr){7-8} \cmidrule(lr){9-10}
			& &FDR &Power &FDR &Power&FDR &Power&FDR &Power\\
			\midrule
			$p=500$ & $\rho=0$ &0.064 &0.910 &0.083 &0.907 &0.060 &0.890 &0.135 &0.907\\
			&  $\rho=0.2$ &0.065 &0.903 &0.071 &0.905 &0.096 &0.887 &0.092 &0.900\\
			& $\rho=0.4$ &0.068 &0.902 &0.012 &0.815 &0.080 &0.898 &0.094 &0.913\\
			& $\rho=0.6$ &0.071 &0.893 &0.024 &0.870 &0.067 &0.900 &0.169 &0.927\\
			& $\rho=0.8$ &0.076 &0.880 &0.008 &0.838 &0.103 &0.882 &0.168 &0.907\\
			\midrule
			$p=2000$ & $\rho=0$  &0.076 &0.798 &0.017 &0.842 &0.053 &0.860 &0.146 &0.787\\
			&  $\rho=0.2$ &0.053 &0.808 &0.058 &0.873 &0.062 &0.850 &0.058 &0.740\\
			& $\rho=0.4$  &0.072 &0.798 &0.014 &0.810 &0.044 &0.840 &0.096 &0.800\\
			& $\rho=0.6$ &0.083 &0.805 &0.048 &0.812 &0.063 &0.810 &0.125 &0.767\\
			& $\rho=0.8$ & 0.063 &0.798 &0.330 &0.815 &0.073 &0.737 &0.177 &0.740\\
			\bottomrule
		\end{tabular}
	\end{threeparttable}
\end{table*}

\section{Additional simulation results with varying correlation}
Consider the linear model $y_i=\bbeta^\top \bx_i+\epsilon_i,  \ \epsilon_i\stackrel{\mbox{\small{i.i.d.}}}{\sim} N(0,1)$,
for $i=1,\dots, n$. We randomly set $k=30$ elements in $\bbeta$ to be nonzero and generated from $\mathcal{N}(0,(20\sqrt{log(p)/n})^2)$ to mimic various signal strengths in real applications. We set $n=1000$ and study both the low-dimensional case with $p=500$ and the high-dimensional case with $p=2000$ respectively.

As shown in the Table \ref{tab:varying_cpc}, S-SNGM and S-INGM have exact FDR control under $q=0.1$ and meanwhile have higher power than the DeepPINK. Besides, SNGM controls FDR under $q=0.1$ except for the case with $p=2000,\rho=0.8$ that is diffcult for mirror construction. In the high-dimensional setting with high partial correlation, DeepPINK undergoes an obvious power loss since the minimum eigenvalue of correlation matrix is $1/(1+(p-1)\rho) \approx 0$, which imposes great obstacles for the Knockoff construction, but NGM with screening procedure is more stable in power maintenance.

\bibliography{main}
\bibliographystyle{plain}
\nocite{*}

\end{document}